%% file: main.tex
\pdfoutput=1

\documentclass[11pt, table]{article}
\usepackage{enumitem}
\usepackage[preprint]{acl}

\usepackage{times}
\usepackage{latexsym}

\usepackage{amsmath}
\usepackage{mathrsfs}
\usepackage{graphicx}
\usepackage{makecell}

\usepackage[T1]{fontenc}

\usepackage[utf8]{inputenc}

\usepackage{microtype}
\usepackage{booktabs}
\usepackage{tabularx}
\usepackage{float}
\usepackage{amsmath}
\usepackage{caption}
\usepackage{subcaption}
\usepackage{algorithm}
\DeclareMathOperator*{\argmax}{arg\,max}

\usepackage{algpseudocode}

\usepackage{tcolorbox}
\usepackage{graphicx}
\usepackage{amsfonts}
\usepackage{xcolor}
\usepackage{subcaption}

\usepackage{bm}
\usepackage{pifont}
\definecolor{cycle2}{RGB}{106, 191, 0}
\definecolor{cycle3}{RGB}{191, 0, 0}
\newcommand{\cmark}{\textcolor{cycle2}{\ding{52}}}
\newcommand{\xmark}{\textcolor{cycle3}{\ding{56}}}

\usepackage{inconsolata}
\usepackage{multirow,tabularx}

\usepackage{listings}

\definecolor{codegreen}{rgb}{0,0.6,0}
\definecolor{codegray}{rgb}{0.5,0.5,0.5}
\definecolor{codepurple}{rgb}{0.58,0,0.82}
\definecolor{backcolour}{rgb}{0.95,0.95,0.92}

\lstdefinestyle{mystyle}{
    backgroundcolor=\color{backcolour},   
    commentstyle=\color{codegreen},
    keywordstyle=\color{magenta},
    numberstyle=\tiny\color{codegray},
    stringstyle=\color{codepurple},
    basicstyle=\ttfamily\footnotesize,
    breakatwhitespace=false,         
    breaklines=true,                 
    captionpos=b,                    
    keepspaces=true,                 
    numbers=left,                    
    numbersep=5pt,                  
    showspaces=false,                
    showstringspaces=false,
    showtabs=false,                  
    tabsize=2
}

\title{Mixture of Structural-and-Textual Retrieval \\over Text-rich Graph Knowledge Bases}

\author{ 
    Yongjia Lei\textsuperscript{1},
    Haoyu Han\textsuperscript{2},
    Ryan A. Rossi\textsuperscript{3},
    Franck Dernoncourt\textsuperscript{3},\\
    \textbf{Nedim Lipka\textsuperscript{3},
    Mahantesh M Halappanavar\textsuperscript{4},
    Jiliang Tang\textsuperscript{2}, 
    Yu Wang\textsuperscript{1}}  \\
    \textsuperscript{1}University of Oregon,
    \textsuperscript{2}Michigan State University,
    \\\textsuperscript{3}Adobe Research,
    \textsuperscript{4}Pacific Northwest National Laboratory \\
    \texttt{\{yuwang, yongjia\}@uoregon.edu, \{hanhaoy1, tangjili\}@msu.edu} \\
    \texttt{\{ryrossi, dernonco, lipka\}@adobe.com, hala@pnnl.gov} 
}

\begin{document}
\maketitle

\begin{abstract}
Text-rich Graph Knowledge Bases (TG-KBs) have become increasingly crucial for answering queries by providing textual and structural knowledge. 
However, current retrieval methods often retrieve these two types of knowledge in isolation without considering their mutual reinforcement and some hybrid methods even bypass structural retrieval entirely after neighboring aggregation. To fill in this gap, we propose a \underline{M}ixture \underline{o}f Structural-and-Textual \underline{R}etrieval (MoR) to retrieve these two types of knowledge via a Planning-Reasoning-Organizing framework. In the Planning stage, MoR generates textual planning graphs delineating the logic for answering queries. Following planning graphs, in the Reasoning stage, MoR interweaves structural traversal and textual matching to obtain candidates from TG-KBs. In the Organizing stage, MoR further ranks fetched candidates by their structural trajectory. Extensive experiments demonstrate the superiority of MoR in harmonizing structural and textual retrieval with discovered insights, including uneven retrieving performance across different query logics and the benefits of integrating structural trajectories for candidate reranking. Our code is available at \href{https://github.com/Yoega/MoR}{https://github.com/Yoega/MoR}.

\end{abstract}

\input{Introduction}

\input{Preliminary}

\input{Method}

\input{Experiment}

\input{RelatedWork}

\input{Conclusion}

\input{Limitations}

\bibliography{reference}

\appendix

\input{Appendix}

\end{document}

%% file: Introduction.tex
\section{Introduction}
Text-rich Graph Knowledge Bases (TG-KBs), due to their structured representation of textual documents, ubiquitously store textual and structural knowledge~\cite{jin2024bridging}. For example, scholars retrieve relevant research from paper management systems to advance scientific discoveries where nodes represent papers and edges denote references~\cite{lu2024ai}. With large language models (LLMs)-powered generators approaching human intelligence in language comprehension and generation, retrieving supporting knowledge from TG-KBs to contextualize and ground generation has become increasingly crucial for correctly answering queries~\cite{gao2023retrieval, ni2025towards}.

\begin{figure}[t!]
    \centering
    \includegraphics[width=0.5\textwidth]{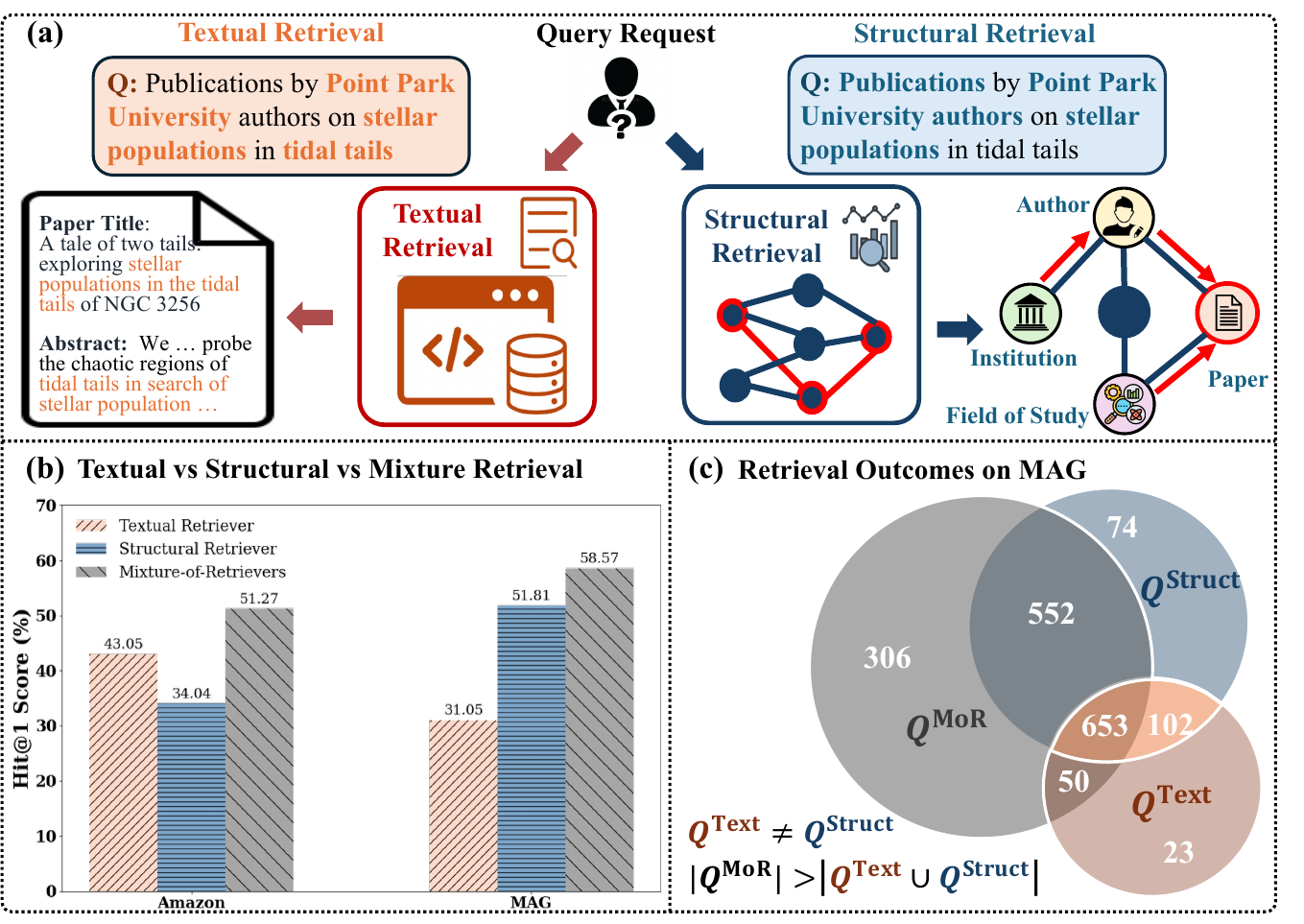}
    \caption{(a) Textual retrieval and structural retrieval. (b) 
    The effectiveness of retrieval methods varies across different TG-KBs. (c) Within the same TG-KB, queries addressed by textual (i.e., $\mathcal{Q}^{\text{Text}}$) and structural retrieval (i.e., $\mathcal{Q}^{\text{Struct}}$) exhibit both overlaps and distinctiveness.}
    \label{fig-motivation}
    \vspace{-3ex}
\end{figure}

Since supporting knowledge in TG-KBs typically exhibits in both the textual and structural formats~\cite{kolomiyets2011survey, jin2024bridging}, retrieval methods should be tailored to both formats effectively as shown in Figure~\ref{fig-motivation}(a). Textual retrieval methods retrieve textual knowledge such as indexed documents based on its similarity to the given query and can be broadly categorized into lexical methods (e.g., BM25) and semantic methods (e.g., Contriever)~\cite{karpukhin-etal-2020-dense, izacardunsupervised}. Structural retrieval methods retrieve structural knowledge such as neighboring entities~\cite{Jiang2023StructGPTAG, Edge2024FromLT, wang2024knowledge} by applying graph traversal and graph machine learning~\cite{yasunaga2021qagnn, tian2024graph}. Despite the advancements in both textual and structural retrieval, they are often applied independently and fail to mutually reinforce each other. As shown by Figure~\ref{fig-motivation}(b), neither structural retrieval by following the logical structure of the query nor textual retrieval by conducting Top-K BM25 matching can achieve better performance on both Amazon and MAG datasets simultaneously.
 
To effectively retrieve both textual and structural knowledge from TG-KBs, recent works~\cite{xia2024knowledge, li2024multi} aggregate neighboring documents to fuse structural knowledge into textual narratives, followed by textual retrieval, with \citet{xia2024knowledge} filtering irrelevant neighbors by their relations and \citet{li2024multi} weighted aggregating neighbors based on their fields. However, three challenges remain. First, rewording aggregated neighbors requires frequently invoking LLMs, resulting in prohibitive resources for long documents with exponentially growing neighbors. Second, structural signals humans use to form logical plans are completely discarded after neighbor aggregation. Third, rigid neighbor aggregation overlooks varying desires for structural and textual knowledge for different queries and TG-KBs. Even within the same TG-KB, such as MAG in Figure~\ref{fig-motivation}(c), queries answered by textual retrieval (i.e., $\mathcal{Q}^{\text{Text}}$) are different from those by structural retrieval (i.e., $\mathcal{Q}^{\text{Struct}}$).

To address the above three challenges, we infuse the mixture-of-expert philosophy into retrieval design and propose a \underline{M}ixture \underline{O}f Structural-and-Textual \underline{R}etrieval (MoR) in Figure~\ref{fig-framework}. MoR begins with a planning module that generates planning graphs to outline query logics and preserve structural signals without rewording aggregated neighbors, overcoming the first and second challenges. Next, MoR interleaves structural traversal and textual matching in the reasoning module, enabling these two retrievals to reinforce each other. Finally, MoR devises a structure-aware reranker in the organization module that adaptively adjusts the importance of retrieved textual/structural knowledge, addressing the third challenge. Via Planning–Reasoning–Organizing, MoR intelligently retrieves structural and textual knowledge based on query logical structure. Our key contributions are:

\begin{itemize}[leftmargin=*]
    \vspace{-0.5ex}
    \item \textbf{Planning via Textual Graph Generation}: 
    We define retrieval planning as generating textual graphs that outline the logical structure, i.e., the plan, for identifying entities relevant to the query.

    \vspace{-0.5ex}
    \item \textbf{Reasoning via Mixture of Structural-and-Textual Traversal}: We devise a mixed traversal by interweaving textual matching and structural traversal to retrieve knowledge following query logical structure depicted by the generated plan.

    \vspace{-0.5ex}
    \item \textbf{Organizing via Structure-aware Rerank}: 
    With candidates obtained from mixed traversal, we propose a Structure-aware Rerank to select Top-K candidates based on their traversal trajectory.
\end{itemize}

%% file: Preliminary.tex
\begin{figure*}[t!]
    \centering
    \includegraphics[width=1\linewidth]{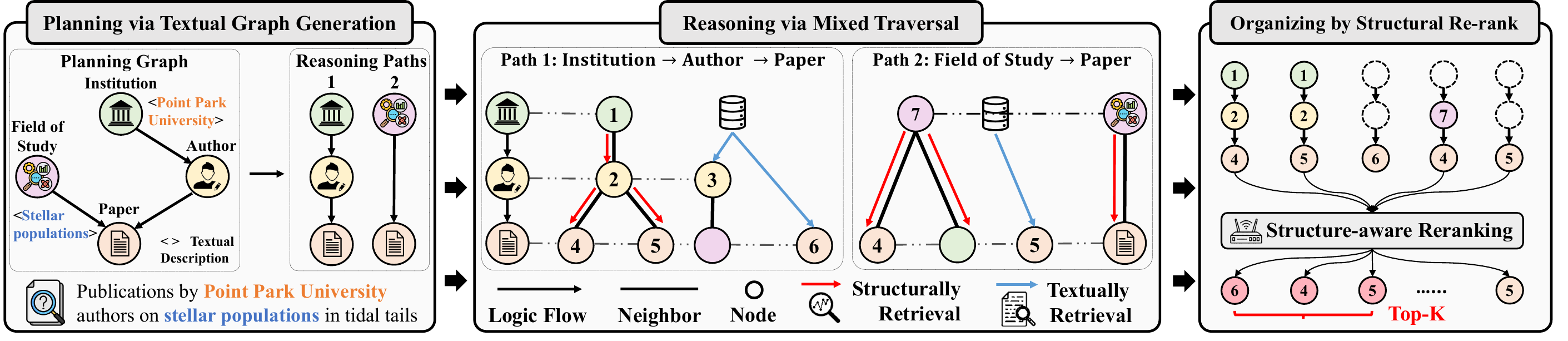}
    \vspace{-4ex}
    \caption{Our MoR framework consists of a planning module to generate a planning graph, a reasoning module to conduct mixed traversal, and an organizing module to rerank the retrieved candidates.}
    \label{fig-framework}
    \vspace{-3ex}
\end{figure*}

\section{Preliminary}\label{sec-preliminary}

\textbf{Notations:} A Text-rich Graph Knowledge Base (TG-KB) $\mathcal{B}$ generally consists of a set of connected nodes $\mathcal{V}$ in the graph with each node $v \in \mathcal{V}$ associated with its corresponding document $\mathcal{D}_v \in \mathcal{D}$ and category $\mathcal{E}_v \in \mathcal{E}$. When retrieving nodes with supporting documents from $\mathcal{B}$ for answering a given query $Q \in \mathcal{Q}$, we typically follow certain rationale encapsulating the underlying logic of that query~\cite{xu-etal-2024-adaption, xue2024enhancing}, which can be characterized by a text-attributed planning graph $G$. In many existing works~\cite{Jin2024GraphCA, wu2024stark}, this planning graph can be usually decomposed into multiple reasoning paths $G = \{\mathcal{P}_i\}_{i=1}^{|G|}$ where the $i^{\text{th}}$ reasoning path $\mathcal{P}_i = (p_{i1} \rightarrow p_{i2} \rightarrow, ..., \rightarrow p_{iL_i})$ is a distinctive reasoning chain of length $L_i$ encoding a unique logic and the $j^{\text{th}}$ node $p_{ij}$ corresponds to an entity in $\mathcal{B}$ with its own category $\mathcal{E}_{p_{ij}}$ and textual restriction $\mathcal{T}_{p_{ij}}$ extracted from the query. For example, in Figure~\ref{fig-motivation}(a), the query \textit{Publications by Point...} has a planning graph with two paths, i.e., $\mathcal{P}_1=(\text{Institution}\rightarrow\text{Author}\rightarrow\text{Paper})$ and $\mathcal{P}_2=(\text{Field-of-Study}\rightarrow\text{Paper})$, where the category and textual restriction of the first node on $\mathcal{P}_1$ are $\mathcal{E}_{p_{11}}=\text{Institution}$ and $\mathcal{T}_{p_{11}}=<\textit{Point Park Univerisity}>$, respectively. Comprehensive notations are summarized in Table~\ref{tab-symbols} in Appendix~\ref{sec-notation}. 

\textbf{Problem Setup:} With the above notations, the investigated problem is to retrieve entities $\mathcal{C} \subseteq \mathcal{V}$ answering a given query $Q$.

\textbf{Textual Retrieval} retrieves candidates based on the textual signals of both the query and documents. One common strategy is to retrieve candidates ${\widetilde{\mathcal{C}}}$ from the whole documents $\mathcal{D}$ that have Top-K textual similarity to query $Q$ measured by lexical or semantic similarity~\cite{vijaymeena2016survey}. The textual retrieval used in MoR retrieves documents for a given query by matching them with textual descriptions in the query, e.g., matching \textit{stellar populations in tidal tails} shown in Figure~\ref{fig-motivation}.

\textbf{Structural Retrieval} retrieves candidates by applying prescribed rules to structured databases such as knowledge graphs and SQL~\cite{guo2023retrieval}. Common strategies include graph-based traversal (e.g., BFS, DFS) and rule fetching~\cite{Jiang2023StructGPTAG}. Specifically, MoR conducts structural retrieval by traversing neighbors of certain categories from the generated planning graph. For example, in Figure~\ref{fig-motivation}(a), only "Paper" typed neighbors of the Author can be traversed by our structural retrieval.

%% file: Method.tex
\section{Framework}\label{sec-framework}
In a nutshell, we formulate our MoR as the conditional distribution $P_{\boldsymbol{\Theta}}(\mathcal{C}|Q, \mathcal{B})$ of retrieved candidates $\mathcal{C}$ given the user input query $Q$ over TG-KB $\mathcal{B}$, which is further factorized into three distributions corresponding to our proposed three modules: planning via generating the text-attributed planning graph $G$, reasoning via conducting mixture of structural-and-textual traversal to obtain intermediate candidates $\widetilde{\mathcal{C}}$ following the generated planning graph $G$, and organizing via applying structure-aware reranking to the obtained candidates $\widetilde{\mathcal{C}}$, obtaining final candidates $\mathcal{C}$:
\begin{equation}
\scalebox{0.9}{$
\begin{aligned}
    &P_{\boldsymbol{\Theta}}(\mathcal{C}|Q, \mathcal{B}) 
    = \sum_{G\in\mathbb{G}} \Bigl[\sum_{\widetilde{\mathcal{C}}\in\mathbb{C}}  
    \underbrace{P_{\boldsymbol{\Theta}_3}(\mathcal{C}|\widetilde{\mathcal{C}}, G, Q, \mathcal{B})}_{\text{Organizing}} \notag \\
    &\quad \times \underbrace{P_{\boldsymbol{\Theta}_2}(\widetilde{\mathcal{C}}|G, Q, \mathcal{B})}_{\text{Reasoning}}\Bigr] 
    \times \underbrace{P_{\boldsymbol{\Theta}_1}(G|Q, \mathcal{B})}_{\text{Planning}}
\end{aligned}
$}
\end{equation}
where $P_{\boldsymbol{\Theta}_1}(G|Q, \mathcal{B})$ is the probability distribution of generating the text-attributed planning graph $G$ given the input query $Q$ and TG-KB $\mathcal{B}$; $P_{\boldsymbol{\Theta}_2}(\widetilde{\mathcal{C}}|G, Q, \mathcal{B})$ is the probability distribution of retrieving intermediate candidates $\widetilde{\mathcal{C}}$ given the planning graph $G$ and the query $Q$ via our mixed traversal; $P_{\boldsymbol{\Theta}_3}(\mathcal{C}|\widetilde{\mathcal{C}}, G, Q, \mathcal{B})$ is the probability distribution of reranking the intermediate candidates so that Top-K positions form the ground-truth entities $\mathcal{C}$. $\mathbb{G}/\mathbb{C}$ denotes the space of all possible planning graphs and all possible configurations of size-K candidate node sets from all nodes $\mathcal{V}$ in TG-KB $\mathcal{B}$. The overall objective is to maximize the likelihood of retrieving ground-truth candidates $\mathcal{C}$ for each input query $Q \in \mathcal{Q}$:
\begin{equation}
\boldsymbol{\Theta}^{*} = \argmax_{\boldsymbol{\Theta}}\prod_{Q \in \mathcal{Q}}P_{\boldsymbol{\Theta}}(\mathcal{C}|Q, \mathcal{B})
\end{equation}

Following the above paradigm, we next introduce the three components: Planning via textual graph generation in Section~\ref{sec-planning}, Reasoning via mixed traversal in Section~\ref{sec-reasoning}, and Organizing via structure-aware reranking in Section~\ref{sec-organizing}.

\subsection{Planning via Textual Graph Generation}\label{sec-planning}

To effectively reason over the underlying logic of queries and answer them, we propose a planning module that constructs a planning graph to capture their underlying logical structures. Unlike conventional approaches relying on rigid heuristics, e.g., shortest-path retrieval in knowledge graphs~\cite{luo2023reasoning, Delile2024GraphBasedRC}, or step-by-step prompting LLMs, which incur high computational costs~\cite{Sun2023ThinkonGraphDA, wang2024knowledge}, our method generates the entire planning graph in one shot, eliminating repeated LLM calls. More importantly, as planning graphs integrate entity restrictions encoding query-specific constraints and entity categories capturing broader logical structure, our MoR can generalize learned patterns and efficiently adapt to new queries with the same underlying logic. For example, any query with the form \textit{Papers associated with <institution> and are in the field of <field>} shares the same patterns with the query in Figure~\ref{fig-framework}. Below, we first formalize the planning graph and then optimize its generation.

\subsubsection{Planning Graph Formulation}
A planning graph $G$ is a structured representation where nodes represent entities and edges denote their logical relations. Each entity is associated with both a category and query-specific restriction. For example, given the query \textit{Can you give me publications by Point Park University authors on stellar populations in tidal tails}, the generated planning graph is: $G=$ (Institution<\textit{Point Park University}> $\rightarrow$ Author $\rightarrow$ Paper $\leftarrow$ Field-of-Study<\textit{Stellar Population}>)  with Institution, Author, Paper, Field-of-Study as categories and <\textit{Point Park University}>, <\textit{Stellar Populations}> as restrictions. Note that edges in our planning graph can also possess different categories. For example, in the biomedical TG-KBs, the relation between Disease and Drug entities could be Indication or Contra-indication~\cite{wu2024stark}, adding a finer level of semantic distinction to the relation.

\subsubsection{Planning Graph Optimization} 
To ensure that our generated planning graph captures the query logic, we train a textual graph generator to maximize the likelihood of generating ground-truth planning graphs given their queries. Formally, given the joint distribution of the training pairs between queries and planning graphs $P_{\mathbb{Q}\times \mathbb{G}}^{\text{Train}}$, we optimize the planning module $P_{\boldsymbol{\Theta}_1}$ by solving:
\begin{equation}
    \arg\max_{\boldsymbol{\Theta}_1} \mathbb{E}_{(Q, G) \sim P_{\mathbb{Q}\times \mathbb{G}}^{\text{Train}}} \log P_{\boldsymbol{\Theta}_1}(G | Q, \mathcal{B}) 
\end{equation}
To avoid the combinatorial explosion of exponentially growing planning graph candidates~\cite{you2018graphrnn}, we decompose each planning graph into multiple reasoning paths $G = \{\mathcal{P}_i\}_{i=1}^{|G|}$. Each path $\mathcal{P}_i = (p_{i1} \rightarrow, ..., \rightarrow p_{iL_i})$ represents a distinct reasoning chain, where node $p_{ij}$ denotes an entity in TG-KB sharing the same textual category $\mathcal{E}_{p_{ij}}$ and satisfying the restriction $\mathcal{T}_{p_{ij}}$ from the query. Given the sequential nature and textual formats of these decomposed reasoning paths, LLMs can be naturally employed here as the planning graph generator, which conducts next-token prediction by predicting $j^{\text{th}}$ token $t_j$ conditioned on preceding tokens $t_{<j}$, the query $Q$ and the TG-KB $\mathcal{B}$:
\begin{equation}
    P_{\boldsymbol{\Theta}_1}(G|Q) = \prod_{j=1}^{n} P_{\boldsymbol{\Theta}_1}(t_j | t_{<j}, Q, \mathcal{B}).
\end{equation}
Note that our proposed planning graph generator is not limited to LLMs. Any graph generative model preserving both structural dependencies and textual associations can be employed~\cite{zhu2022survey}.

\subsection{Reasoning via Mixed Traversal}\label{sec-reasoning}
Following the reasoning paths of the above planning graph $G = \{\mathcal{P}_i\}_{i=1}^{|G|}$, the reasoning module conducts a mixed traversal by interweaving neighbor fetching and textual matching to form intermediate candidates $\widetilde{\mathcal{C}}$, which are introduced next.

\subsubsection{Structural Traversal}
Following the definition in Section~\ref{sec-preliminary} that structural retrieval follows prescribed rules for knowledge retrieval, here we set these prescribed rules to be iteratively performing layer-wise breadth-first-search that traverses neighboring entities with categories aligning with those in the reasoning paths. Concretely, reasoning at the $l^{\text{th}}$-step of the planning path $\mathcal{P}_i$, we check for each node $v$ in candidates set of last layer $\forall v \in \widetilde{\mathcal{C}}_i^{l-1}$ and fetch its neighbors $\forall u \in \mathcal{N}_v$ with the same category as the corresponding node $p_{il}$ (i.e., $\mathcal{E}_u = \mathcal{E}_{p_{il}}$) in the reasoning path, which can be mathematically formulated as:
\begin{equation}
\widetilde{\mathcal{C}}_i^{l, \text{Struct}} = \cup_{v \in \widetilde{\mathcal{C}}_i^{l-1}}\{u|u \in \mathcal{N}_v, \mathcal{E}_u = \mathcal{E}_{p_{il}}\} 
\end{equation}
where $\widetilde{\mathcal{C}}_i^{l, \text{Struct}}$ denotes the set of structurally retrieved entities at the $l^{\text{th}}$ reasoning step according to the path $\mathcal{P}_i$ and $\mathcal{E}_u=\mathcal{E}_{p_{il}}$ ensures that the category of the traversed neighbor $u$ matches the corresponding entity category routine by the planning graph, resonating the nature of rule-based structural retrieval. Note that the seeding candidates $\widetilde{\mathcal{C}}^{1, \text{Struct}}_i$ at the very first layer are initialized by retrieving Top-K entities through textual matching, i.e., $\widetilde{\mathcal{C}}^{1, \text{Struct}}_i = \widetilde{\mathcal{C}}^{1, \text{Text}}_i$, which is introduced next.

\subsubsection{Textual Matching}
In addition to retrieving structural knowledge, our MoR also retrieves textual knowledge via Textual Matching, which retrieves candidates based on their textual similarity to queries. For each reasoning node $p_{il}$ at $l^{\text{th}}$ reasoning step along the reasoning path $\mathcal{P}_i$, we concatenate the query and the textual restriction of $p_{il}$, i.e., $Q' = [Q:\mathcal{T}_{p_{il}}]$, then compute its textual similarity to documents of nodes in TG-KB, i.e., $\phi(Q', \mathcal{D}_v), \forall v \in \mathcal{V}$, and finally retrieve the Top-K scored candidates:
\begin{equation}
\scalebox{0.86}{$
    \widetilde{\mathcal{C}}_i^{l, \text{Text}} = \operatorname{TopK} ( \{ v \mid v \in \mathcal{V}, \mathcal{E}_v = \mathcal{E}_{p_{il}} \}, \phi(Q', \mathcal{D}_v))
$}
\end{equation}

Integrating candidates from structural traversal and textual matching together, the final candidates at $l^{\text{th}}$-step of $\mathcal{P}_i$ are formed as:
\begin{equation}
\scalebox{0.9}{$
    \widetilde{\mathcal{C}}^{l}_i = \widetilde{\mathcal{C}}^{l, \text{Struct}}_i \cup \widetilde{\mathcal{C}}^{l, \text{Text}}_i, \forall l\in\{1, 2, ..., L_i\}
$}
\end{equation}
The integrated candidates $\widetilde{\mathcal{C}}^{l}_i$ serve as seeding nodes initializing the next round of planning graph-guided structural traversal and textual matching, which creates a mutual reinforcement between structural and textual knowledge since previously retrieved two knowledge can both inform next round of structural/textual knowledge retrieval.

We iteratively conduct mixed traversal for every reasoning path $\mathcal{P}_i\in G$ and integrate retrieved entities together by taking their intersection, i.e., $\widetilde{\mathcal{C}}=\cap_{\mathcal{P}_i\in G}\widetilde{\mathcal{C}}_i^{L_i}$, adhering to the fact that candidates should simultaneously satisfy the logic routine by all reasoning paths. Note that no training is involved in the mixed graph traversal, i.e., $P_{\boldsymbol{\Theta}_2}(\widetilde{\mathcal{C}}|G, Q, \mathcal{B}) = P(\widetilde{\mathcal{C}}|G, Q, \mathcal{B})$. Future works could explore optimizing graph traversal by rewards from agent-environment interactions~\cite{nguyen2024dynasaur}.

\newpage
\subsection{Organizing via Structure-aware Rerank}\label{sec-organizing}
Although the retrieved candidates from Section~\ref{sec-reasoning} strictly adhere to the prescribed rule given by the planning graph, the sheer volume of candidates misaligns with realistic constraints (e.g., Top-20 retrieval budget~\cite{Zeng2024FederatedRV}) and may even cause difficulty to downstream executors such as long-context challenges for LLMs. To better emulate human reasoning, where multiple clues are gathered, analyzed in relation to the query, and synthesized into a coherent answer, we propose a structure-aware reranker to organize and rerank the candidates $\widetilde{\mathcal{C}}$, and select Top-K ones as the final retrieved answers $\mathcal{C}$. Instead of relying only on textual features~\cite{hu2019retrieve}, our reranker assigns a ranking score based on features of structural trajectories obtained from the mixed traversal in Section~\ref{sec-reasoning}, innovatively leveraging both structural and textual knowledge in reranking.  

Previously, $\widetilde{\mathcal{C}}$ is defined as intermediate retrieved entities. To consider structural features in reranking, we pair each retrieved candidate in $\widetilde{\mathcal{C}}$ with its corresponding traversal trajectory obtained from the reasoning module. Specifically, each trajectory $\mathcal{P}_i$ of length $L_i$ is featuring three types of attributes:

\begin{itemize}[leftmargin=*]
    \item \textbf{Textual Fingerprint (TF)}: Concatenation of similarity scores between the expanded query and each node on the path: $\big\|_{l=1}^{L_i} \phi(Q', \mathcal{D}_{p_{il}})$.
    \item \textbf{Structural Fingerprint (SF)}: Concatenation of node categories at each step on the path: $\big\|_{l=1}^{L_i} \mathcal{E}_{p_{il}}$
    \item \textbf{Traversal Identifier (TI)}: Concatenation of the indicator specifying whether each step uses a structural or textual retrieval: $\big\|_{l=1}^{L_i} \mathcal{I}_{p_{il}}$.
\end{itemize}
We then train a reranker on these trajectories using the cross-entropy loss. For a training query $Q$ and its associated candidate trajectory $\mathcal{P}_i$, the loss is computed as follows:
\begin{equation}
\scalebox{0.85}{$
\begin{aligned}
    \mathcal{L}_{\boldsymbol{\Theta}_3} =-\sum_{(\mathcal{P}_i, Q) \in \widetilde{\mathcal{C}}}\sum_{j=1}^2 y^i_j\log ( \sigma(&f(\underbrace{\big\|_{l=1}^{L_i}\mathcal{E}_{p_{il}}}_{\text{Structural Fingerprint}}\\    :\underbrace{\big\|_{l=1}^{L_i}\phi(Q', \mathcal{D}_{p_{il}})}_{\text{Textual Fingerprint}} &:\underbrace{\big\|_{l=1}^{L_i}\mathcal{I}_{p_{il}}}_{\text{Traversal Identifier}}))_j).
\end{aligned}
$}
\end{equation}
where $f(\cdot)$ is the reranker producing a score for each $(Q, \mathcal{P}_i)$ pair, $\sigma(\cdot)$ denotes the softmax function, and $y^i_j \in \{0,1\}$ indicates whether the $i$-th candidate is a correct (positive) or incorrect (negative) match for $Q$. This formulation encourages the reranker to assign higher scores to positive trajectories, thereby improving ranking performance.

\begin{table*}[t!]
\setlength\tabcolsep{2.6pt}
\centering
\tiny
\begin{tabular}{ll|cccc|cccc|cccc|cccc}
\toprule
\multirow{3}{*}{\textbf{Category}} & \multirow{3}{*}{\textbf{Retrieval Baseline}}  & \multicolumn{4}{c|}{\textbf{AMAZON}} & \multicolumn{4}{c|}{\textbf{MAG}} & \multicolumn{4}{c|}{\textbf{PRIME}} & \multicolumn{4}{c}{\textbf{AVERAGE}} \\
\cmidrule(lr){3-6} \cmidrule(lr){7-10} \cmidrule(lr){11-14} \cmidrule(lr){15-18}
& & H@1 & H@5 & R@20 & MRR & H@1 & H@5 & R@20 & MRR & H@1 & H@5 & R@20 & MRR & H@1 & H@5 & R@20 & MRR \\
\midrule
\multirow{4}{*}{\textbf{Textual}}       & BM25~\cite{wu2024stark}             & 44.94 & 67.42 & 53.77 & 55.30 & 25.85 & 45.25 & 45.69 & 34.91 & 12.75 & 27.92 & 31.25 & 19.84 & 27.85  & 46.86 & 43.57 & 36.68\\
    & Ada-002~\cite{wu2024stark}         & 39.16 & 62.73 & 53.29 & 50.35 & 29.08 & 49.61 & 48.36 & 38.62 & 12.63 & 31.49 & 36.00 & 21.41 & 26.96 & 47.94 & 45.88 & 36.79\\
    & Multi-ada-002~\cite{wu2024stark}   & 40.07 & 64.98 & 55.12 & 51.55 & 25.92 & 50.43 & 50.80 & 36.94 & 15.10 & 33.56 & 38.05 & 23.49 & 27.03 & 49.66 & 47.99 & 37.33\\
    & DPR~\cite{karpukhin-etal-2020-dense}             & 15.29 & 47.93 & 44.49 & 30.20 & 10.51 & 35.23 & 42.11 & 21.34 & 4.46  & 21.85 & 30.13 & 12.38 & 10.09 & 35.00 & 38.91 & 21.31\\
\midrule
\multirow{2}{*}{\textbf{Structural (KG)}} & QAGNN~\cite{yasunaga2021qagnn}       & 26.56 & 50.01 & 52.05 & 37.75 & 12.88 & 39.01 & 46.97 & 29.12 & 8.85  & 21.35 & 29.63 & 14.73 & 16.10 & 36.79 & 42.88     & 27.20\\
& ToG~\cite{Sun2023ThinkonGraphDA} & - & - & - & - & 13.16 & 16.17 & 11.30 & 14.18 & 6.07  & 15.71 & 13.07 & 10.17 & 9.62 & 15.94    & 12.18 & 12.18\\
\midrule
\multirow{4}{*}{\textbf{Hybrid}} & AvaTaR~\cite{wu2024avatar}  & 49.87 & 69.16 & \textbf{60.57} & 58.70 & 44.36 & 59.66 & 50.63 & 51.15 & 18.44 & 36.73 & 39.31 & 26.73  & 37.56 & 55.18 & 50.17 & 45.53\\
& KAR~\cite{xia2024knowledge}             & \textbf{54.20} & 68.70 & 57.24 & \underline{61.29} & 50.47 & 69.57 & 60.28 & 58.65 & 30.35 & 49.30 & 50.81 & 39.22 & 45.01 & 62.52  & 56.11 &53.05\\

& MFAR$^{*}$~\cite{li2024multi} & 41.20 & \underline{70.00} & 58.50 & 54.20 & 49.00 & \underline{69.60} & \underline{71.70} & 58.20 & \textbf{40.90} & \textbf{62.80} & \textbf{68.30} & \textbf{51.20} & 43.70 & \underline{67.47} & \textbf{66.17} & \underline{54.53}\\
& HYBGRAG~\cite{Lee2024HybGRAGHR} &  - &  - & - & - 
& \textbf{65.40} & 75.31 & 65.70 & \textbf{69.80} 
&  28.56 & 41.38 & 43.58 & 34.49 
& \textbf{50.91} & 58.35 & 54.64 & 52.15 \\

& \cellcolor{gray!30} MoR & \cellcolor{gray!30} \underline{52.19} & \cellcolor{gray!30} \textbf{74.65} & \cellcolor{gray!30} \underline{59.92} & \cellcolor{gray!30} \textbf{62.24} & \cellcolor{gray!30} \underline{58.19} & \cellcolor{gray!30} \textbf{78.34} & \cellcolor{gray!30} \textbf{75.01} & \cellcolor{gray!30} \underline{67.14} & \cellcolor{gray!30} \underline{36.41} & \cellcolor{gray!30} \underline{60.01} & \cellcolor{gray!30} \underline{63.48} & \cellcolor{gray!30} \underline{46.92} & \cellcolor{gray!30} \underline{48.93} & \cellcolor{gray!30} \textbf{71.00}  & \cellcolor{gray!30} \underline{66.14} & \cellcolor{gray!30} \textbf{58.77}\\
\midrule
\multirow{3}{*}{\textbf{Ablation}} &\cellcolor{gray!30} MoR$_{\text{w/o R}}$ &\cellcolor{gray!30} 44.21  &\cellcolor{gray!30} 68.87 &\cellcolor{gray!30} 56.50 &\cellcolor{gray!30} 55.28 &\cellcolor{gray!30} 34.33 &\cellcolor{gray!30} 62.55 &\cellcolor{gray!30} 67.55 &\cellcolor{gray!30} 47.40 &\cellcolor{gray!30} 31.59 
&\cellcolor{gray!30} 53.48 &\cellcolor{gray!30} 60.74 & \cellcolor{gray!30} 41.81 &\cellcolor{gray!30} 31.07 &\cellcolor{gray!30} 57.04 &\cellcolor{gray!30} 57.73 &\cellcolor{gray!30} 43.03\\

&\cellcolor{gray!30} MoR$_{\text{w/o RT}}$& \cellcolor{gray!30} 34.04  &\cellcolor{gray!30} 53.41 &\cellcolor{gray!30} 45.16 &\cellcolor{gray!30} 42.85 &\cellcolor{gray!30} 51.81 &\cellcolor{gray!30} 73.54 &\cellcolor{gray!30} 74.17 &\cellcolor{gray!30} 61.68 &\cellcolor{gray!30} 28.95 &\cellcolor{gray!30} 46.12 &\cellcolor{gray!30} 49.54 &\cellcolor{gray!30} 36.56 &\cellcolor{gray!30} 36.39 &\cellcolor{gray!30} 56.73 &\cellcolor{gray!30} 55.73 &\cellcolor{gray!30} 45.53 \\

&\cellcolor{gray!30} MoR$_\text{w/o RS}$ &\cellcolor{gray!30} 43.05  &\cellcolor{gray!30} 69.36 &\cellcolor{gray!30} 57.38 &\cellcolor{gray!30} 54.69 &\cellcolor{gray!30} 31.05 &\cellcolor{gray!30} 51.84 &\cellcolor{gray!30} 50.56 &\cellcolor{gray!30} 40.64 &\cellcolor{gray!30} 22.27 &\cellcolor{gray!30} 38.45 &\cellcolor{gray!30} 39.21 &\cellcolor{gray!30} 29.41 &\cellcolor{gray!30} 28.95 &\cellcolor{gray!30} 51.28 &\cellcolor{gray!30} 48.02 &\cellcolor{gray!30} 38.98\\

\bottomrule
\end{tabular}
\vspace{-3ex}
\caption{Comparing different retrieval methods with our proposed MoR and its ablations on Amazon, MAG, and Prime datasets. The best and runner-up results are in \textbf{bold} and \underline{underlined}. Overall, MoR achieves the best performance. Note that MFAR$^{*}$ denotes the best model variant proposed in~\cite{li2024multi}}
\label{tab-merged}
\vspace{-3ex}
\end{table*}

%% file: Experiment.tex
\newpage
\section{Experiment}\label{sec-experiment}
\subsection{Experimental Setup}
We briefly introduce experimental settings to verify our proposed MoR, including Datasets \& Baselines, Implementation Details, and Evaluation Metrics. More details are in Appendix~\ref{app-expr-setting}.

\textbf{Datasets \& Baselines:} We use three TG-KBs from STaRK~\cite{wu2024stark} covering three knowledge domains, including E-commerce Products (Amazon), Academic Papers (MAG), and Biomedicine (Prime). We compare our MoR with baselines established by~\citet{wu2024stark} and categorize them into textual/structural/hybrid-based ones. More recent state-of-the-art hybird retrieval approaches fro TG-KBs such as KAR~\cite{xia2024knowledge} and MFAR$^{*}$~\cite{li2024multi} are also compared.

\textbf{Implementation Details:} 
To enhance the planning capability of our planning module, we fine-tune the Llama 3.2 (3B) on 1000 sampled queries with their corresponding ground-truth planning graphs, serving as the textual graph generator. In the absence of ground-truths, we synthesize them using LLMs. For the Prime dataset, we empirically find that directly prompting LLMs can hardly generate accurate planning graphs due to the lack of biomedical domain knowledge~\cite{Shen2024TagLLMRG}. Therefore, we adopt an alternative approach. First, we instruct LLMs to extract triplets from each query and then construct the planning graphs by merging triplets with shared entities. 
During mixed traversal, textual matching can be implemented using any lexical or semantic methods. For this study, we employ BM25 for Amazon and MAG and fine-tune a contriever to complement the biomedical knowledge for Prime.
To initialize the structural traversal, we employ textual matching to locate the top 5 nodes that are most relevant to the query as seeds. Additionally, at each layer, we incorporate the top 10 nodes retrieved via textual matching and append them to the current candidate set for the next round of traversal. Notably, due to the uncertainty of LLMs, the generated planning graphs can be invalid. In this case, we will directly conduct textual matching to retrieve candidates. For our ablations without reranker, we employ Ada-002~\cite{wu2024stark} with cosine similarity as the scorer to rank candidates for evaluating performance.

\textbf{Evaluation Metrics:}
We follow~\citet{wu2024stark} for evaluation by reporting Hit@1 (H@1), Hit@5 (H@5), Recall@20 (R@20), and mean reciprocal rank MRR to evaluate in the full spectrum.

\newpage
\subsection{Overall Retrieval Performance}
We compare MoR with other baselines on three TG-KBs in Table~\ref{tab-merged}. Generally, hybrid methods, AvaTAR, KAR, MFAR$^{*}$, and our MoR, achieve better performance than purely textual or structural methods owing to their ability to integrate both structural and textual knowledge. 
Among all baselines, our proposed MoR achieves the overall best performance with a substantial margin on average, with the first ranking on MAG and the second ranking on Amazon/Prime datasets. This demonstrates the effectiveness of our proposed mixture of structural and textual knowledge retrieval. 
Textual retrieval performs better on Amazon than on MAG, suggesting that Amazon queries rely more on textual knowledge. In contrast, its weaker performance on MAG is due to MAG's lower textual richness and stronger structural signals. This disparity aligns with the distribution analysis presented by~\citet{wu2024stark} and supports our hypothesis that queries in different TG-KB datasets require varying desires for textual and structural knowledge. Meanwhile, structural retrieval methods such as conventional knowledge graph-based ones perform poorly because they are designed for graphs with minimal textual information compared to TG-KBs.
Different from Amazon and MAG, all existing methods without supervised tuning (e.g., Ada-002) exhibit significantly lower performance on Prime. This is due to the extreme domain expertise required in biology, where word-count-based, pre-trained textual similarity-based, and even more powerful LLMs are all poorly applicable here. Through fine-tuning, MFAR$^{*}$ and our proposed MoR generally achieve better performance, demonstrating the necessity of domain-specific knowledge for answering queries in knowledge-intensive domains.

\newpage
\subsection{Ablation Study}
After verifying the superiority of MoR, we conduct ablation studies to assess its different components, including module and feature ablation.

\subsubsection{Module Ablation}

To assess the contribution of each module in MoR, namely, Text Matching-based Retrieval, Neighborhood-Fetching-based Structural Retrieval, and Reranker, we conduct a series of ablation experiments. First, we remove the Reranker, resulting in the variant MoR$_{\text{w/o R}}$. On top of that, we further separately eliminate Text Retrieval and Structural Retrieval, yielding MoR$_{\text{w/o RT}}$ and MoR$_{\text{w/o RS}}$, respectively.
As shown in Table~\ref{tab-merged}, the complete MoR framework consistently achieves the highest performance across all datasets, demonstrating the synergistic effect of the Textual Retriever, Structural Retriever, and Reranker.
After removing Reranker, MoR$_{\text{w/o R}}$ exhibits a consistent performance drop across all datasets and evaluation metrics. This underscores the importance of the Reranker in refining retrieval by filtering noisy candidates from the intermediate reasoning stage. 
Eliminating Text Retrieval, i.e., MoR$_{\text{w/o RT}}$, leads to a notable performance drop on Amazon but an unexpected improvement on MAG. This suggests that while textual knowledge benefits Amazon, it introduces misleading hard negatives that compromise the ranking method (e.g., Ada-002) for MAG. Conversely, removing Structural Retrieval, MoR$_{\text{w/o RS}}$, results in a slight performance decrease further on MAG, reinforcing the importance of structural knowledge in MAG-related queries.
These results underscore the Reranker's crucial role in adaptively harmonizing, balancing, and selecting knowledge from both structural and textual retrieval experts.

\begin{table}[t!]

\setlength\tabcolsep{6pt}
\centering
\tiny
\resizebox{0.49\textwidth}{3cm}{
\begin{tabular}{l|ccc|cccc}
\toprule
\textbf{Dataset} &\textbf{TF} & \textbf{SF} & \textbf{TI} & \textbf{H@1} & \textbf{H@5} & \textbf{R@20} & \textbf{MRR} \\ \midrule
\multirow{7}{*}{\textbf{MAG}} 
& \cmark & \xmark & \xmark & 48.96 & 73.02 & 72.44 & 59.79 \\
&      \xmark            & \cmark       &         \xmark         & 18.79 & 41.91 & 52.85 & 29.84 \\
&        \xmark          &         \xmark         & \cmark       & 18.16 & 41.53 & 52.78 & 29.31 \\
\cline{2-8}
& \cmark       & \cmark       &    \xmark              & 58.04 & 77.14 & 74.42 & 66.75 \\
& \cmark       &        \xmark          & \cmark       & \underline{58.16} & \underline{77.59} & \underline{74.96} & \underline{66.85} \\
&          \xmark        & \cmark       & \cmark       & 17.93 & 38.01 & 46.79 & 27.48 \\
\cline{2-8}
& \cmark       & \cmark       & \cmark       & \textbf{58.19} & \textbf{78.34} & \textbf{75.01} & \textbf{67.14} \\ \midrule
\multirow{7}{*}{\textbf{Amazon}}    
& \cmark       &      \xmark            &       \xmark           & \underline{51.21} & \underline{74.05} & \underline{59.79} & \underline{61.27} \\
&        \xmark          & \cmark       &      \xmark            & 8.09  & 24.48 & 25.62 & 16.94 \\
&         \xmark         &      \xmark            & \cmark       & 5.84  & 16.62 & 12.94 & 11.57 \\
\cline{2-8}
& \cmark       & \cmark       &      \xmark            & 50.91 & 73.38 & 59.58 & 61.15 \\
& \cmark       &         \xmark         & \cmark       & 51.09 & 73.56 & 59.61 & 61.14 \\
&            \xmark      & \cmark       & \cmark       & 8.09  & 24.48 & 25.62 & 16.94 \\
\cline{2-8}
& \cmark       & \cmark       & \cmark       & \textbf{52.19} & \textbf{74.65} & \textbf{59.92} & \textbf{62.24} \\ 
\midrule
\multirow{7}{*}{\textbf{Prime}}    
& \cmark & \xmark & \xmark & 35.23 & 59.44 & 63.15 & 46.02 \\
& \xmark & \cmark & \xmark & 12.95 & 30.48 & 43.33 & 21.44 \\
& \xmark & \xmark & \cmark & 11.81 & 27.81 & 40.36 & 19.73 \\
\cline{2-8}
& \cmark & \xmark & \cmark & \underline{35.80} & \textbf{60.12} & 63.40 & \underline{46.50} \\
& \cmark & \cmark & \xmark & \underline{35.70} & \underline{60.01} & \underline{63.21} & \underline{46.19} \\
& \xmark & \cmark & \cmark & 13.20 & 32.27 & 48.01 & 22.95 \\
\cline{2-8}
& \cmark & \cmark & \cmark & \textbf{36.41} & \underline{60.01} & \textbf{63.48} & \textbf{46.92} \\
\bottomrule
\end{tabular}
}
\caption{Ablation study investigating the importance of three features, Textual Fingerprint (\textbf{TF}), Structural Fingerprint (\textbf{SF}), and Traversal Identifier (\textbf{TI}), of the traversal trajectories used in our Structure-aware Reranker.}
\label{tab-feature-ablation}
\vspace{-4ex}
\end{table}

\subsubsection{Feature Ablation}
The above ablation study highlights the crucial role of Structure-aware Reranker in adaptively integrating structural and textual knowledge. To further analyze the contributions of its three key features, \textbf{Textual Fingerprint (TF)}, \textbf{Structural Fingerprint (SF)}, and \textbf{Traversal Identifier (TI)} defined in Section~\ref{sec-organizing}, we conduct a feature ablation analysis and report retrieval performance across different feature configurations in Table~\ref{tab-feature-ablation}.
Overall, using three features together yields the best performance on both MAG and Amazon, highlighting their synergistic effect. Individually, TF contributes the most and outperforms SF and TI on both datasets. 
The reason is that based on the definition in Section~\ref{sec-organizing}, TF directly captures the relevance between the query and the retrieved nodes along the trajectory, whereas SF and TI primarily characterize the structural patterns and retrieval types, serving more as complementary factors. Therefore, equipping TF with these complementary factors (i.e., SF or TI) yields around 10\% additional gains on MAG. This is because SF and TI help the reranker selectively emphasize the relevance scores given by TF for certain nodes along the path. However, this boost is not observed on Amazon. We hypothesize that the textual knowledge needed there is predominantly derived from the final node on each path, making the structural cues provided by SF and TI less beneficial and even prone to overfitting. A deeper analysis to further justify this hypothesis is in Section~\ref{sec-further}. Overall, these findings underscore the varying importance of structural features in ranking across datasets.

\begin{table}[t!]
\setlength\tabcolsep{2pt}
\centering
\resizebox{0.48\textwidth}{!}{
\begin{tabular}{l|ccc|ccc|ccc}
\toprule
\multirow{2}{*}{\textbf{Feature}} & \multicolumn{3}{c|}{\textbf{MAG}} & \multicolumn{3}{c}{\textbf{Amazon}} & \multicolumn{3}{c}{\textbf{Prime}} \\

 & H@1 & R@20 & MRR & H@1 & R@20 & MRR & H@1 & R@20 & MRR \\
\midrule
Last Node & 49.91 & 73.49 & 59.92 & 50.36 & 59.62 & 61.05 & 33.52 & 61.95 & 44.15   \\
Full Path & \textbf{58.19} & \textbf{75.01} & \textbf{67.14} & \textbf{52.19} & \textbf{59.92} & \textbf{62.24} & \textbf{36.41} & \textbf{63.48} & \textbf{46.92}   \\
\bottomrule
\end{tabular}
}
\caption{Comparing reranking performance using last node in the retrieved trajectory and the whole trajectory.}
\label{tab-Reranker-ablation}
\vspace{-2ex}
\end{table}

\begin{figure}[t!]
    \centering
    \includegraphics[width=0.49\textwidth, height = 0.22\textwidth]{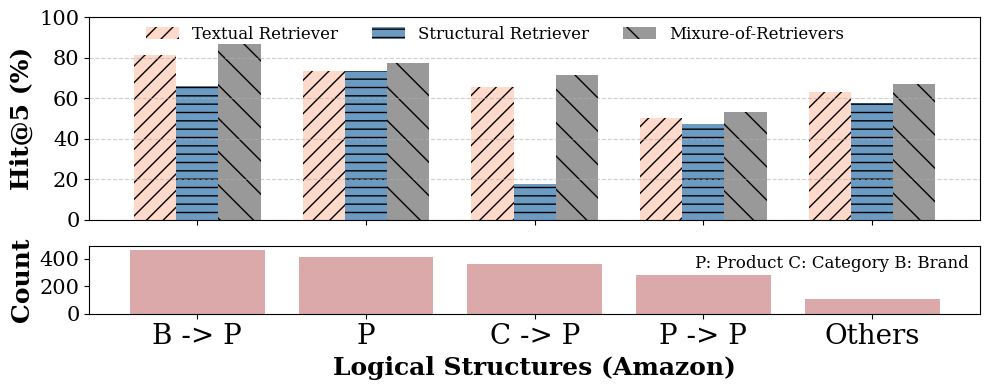}
    \vspace{-4.5ex}
    \caption{Imbalance number of queries and performance of different retrievers across different logical structures.}
    \label{fig-analysis}
    \vspace{-3ex}
\end{figure}

\subsection{Further Analysis}\label{sec-further}
This section understands MoR’s behavior by examining three questions, each of which enriches our insight into MoR’s functionality and offers novel perspectives inspiring future query retrieval research. More analysis can be found in Appendix~\ref{app-results}.

\textbf{Do structure signals affect reranking?}
To assess the impact of trajectory information on the Reranker's decision-making, we introduce a node-based Reranker that constructs trajectory features using only TF/SF/TI of the last node. In Table~\ref{tab-Reranker-ablation}, the path-based Reranker outperforms the node-based variant, especially on MAG. This highlights the critical role of trajectory features/structural knowledge in reranking. The minor performance boost on Amazon after switching to the full path trajectory indicates its textual knowledge preference over the last node rather than the whole trajectory.

\textbf{How does MoR perform on different logical structures?}
Figure~\ref{fig-analysis} shows the average performance of MoR on each query group categorized by their logical structures, where "Others" refer to queries with undefined logical structures in~\citet{wu2024stark} MoR consistently outperforms structural and textual retrievers across different logical structures. Among all queries, MoR performs the worst on "P → P" queries due to the ambiguity uniquely caused by repeated "product" entities from multi-step traversal.
The average-performing ``Others" group underscores the utility of diverse planning strategies for the same query.
Lastly, the skewed query distribution and retrieval performance across planning patterns reflect the varying nature of real-world planning needs. We hope these insights inspire data-centric reasoning and error control of planning for heterogeneous query structures.

\begin{figure}[t!]
    \centering
    \includegraphics[width=0.5\textwidth]{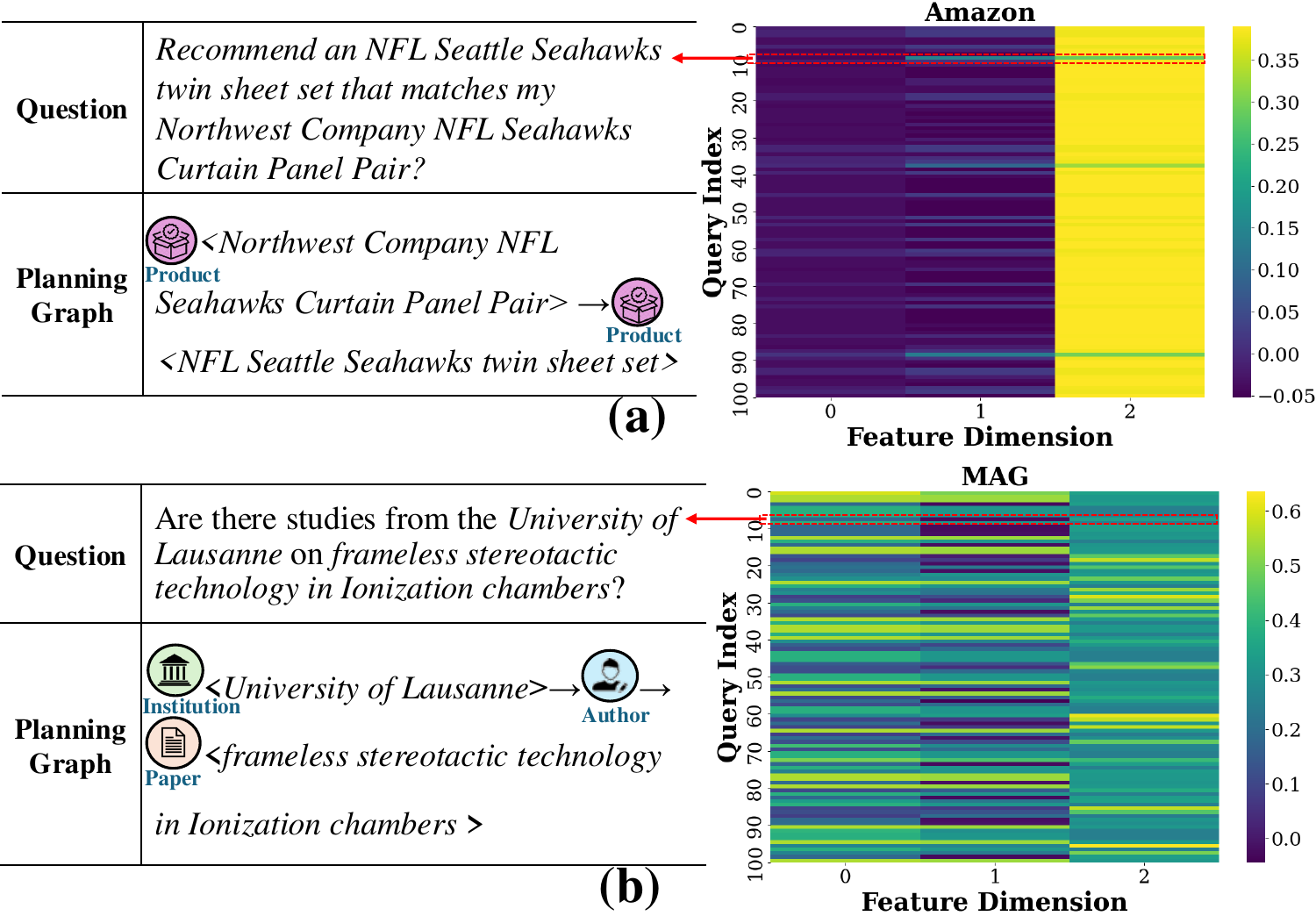}
    \vspace{-3ex}
    \caption{Saliency map visualization of query attention over three entities along the retrieved paths.}
    \label{fig-map}
    \vspace{-4ex}
\end{figure}

\textbf{Does MoR indeed adaptively leverage the trajectory knowledge?} To understand how our proposed reranker prioritizes candidates in the Top-K results, we visualize the saliency map by computing the gradient of ranking scores with respect to the textual fingerprint (TF) of three nodes along the traversed path, which quantifies their importance for answering a given query. Figure~\ref{fig-map} illustrates this by analyzing trajectories for 100 ground-truth candidates across 100 queries on the Amazon and MAG datasets. Each dimension corresponds to a traversed node, with the final one representing the candidate itself. 
While the saliency score is concentrated in the last dimension for Amazon, 
MAG exhibits a more evenly distributed saliency pattern, where multiple nodes along the path contribute significantly to ranking score computation. This suggests that structural knowledge is more critical for answering queries in MAG, aligning with the previously observed lower performance of purely textual retrieval on MAG in Table~\ref{tab-merged}. Further case studies explain why the reranker attends different nodes for different queries. In Figure~\ref{fig-map}(a), the reranker favors the last two dimensions as the rich textual restriction (i.e., "Northwest Company..." and "NFL Seattle...") aids in identifying the correct node at the corresponding reasoning step, as discussed in Section~\ref{sec-reasoning}. These correct nodes with higher similarity scores with the query help guide the retrieval process toward the ground truth.
Conversely, in Figure~\ref{fig-map}(b),
since the first node ("University of Lausanne") helps narrow the search space and the last node ("frameless...") further filter candidates, both nodes have high saliency scores. Overall, our findings demonstrate that the reranker dynamically adapts its reliance on structural and textual knowledge depending on the dataset and query.

\begin{figure}[t!]
    \centering
    \includegraphics[width=0.5\textwidth]{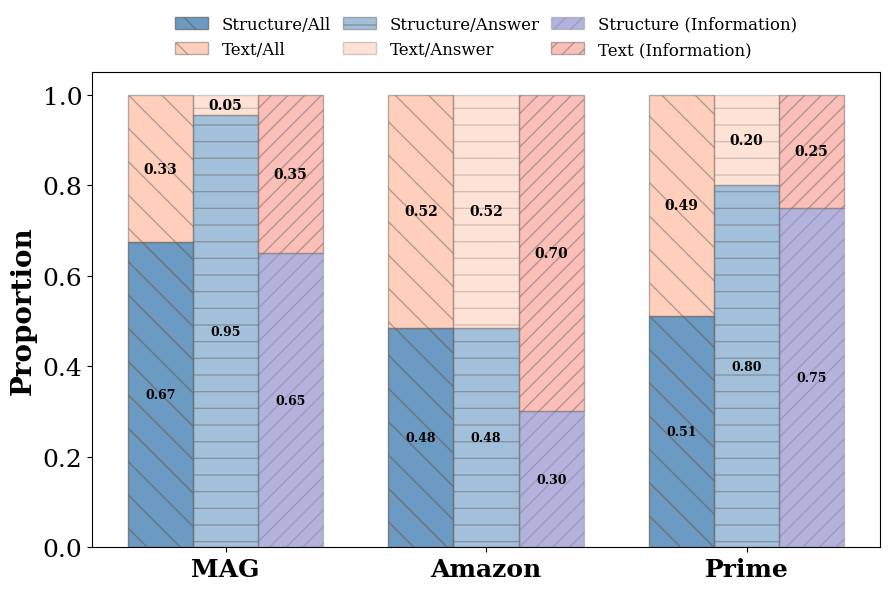}
    \caption{Traversal Identifier analysis in Top-20 retrieved candidates and information ratio in dataset.}
    \label{fig-top20}
\end{figure}

\textbf{Does MoR adaptively prioritize retrieval methods for different datasets?} 
To assess the degree to which MoR adaptively leverages structural and textual retrieval across datasets, we calculate the following three ratios: \textbf{Structure/All, Text/All:} Fraction of structurally or textually retrieved candidates within the Top-20 retrieved candidates; \textbf{Structure/Answer, Text/Answer:} Fraction of ground-truth candidates that are structurally or textually retrieved among the Top-20; \textbf{Structure (Information), Text (Information)}: Ratio of word counts of sampled relation/property requirements used in query construction.


The results are shown in Figure~\ref{fig-top20}. On the MAG dataset, a majority of the Top-20 retrieved candidates (67\%) and over 95\% of the correct answers originate from structural retrieval. This underscores the critical role of structural signals in this domain. The finding is consistent with the inherently rich structural nature of MAG, evidenced by the higher proportion of \textbf{Structure (Information)} compared to \textbf{Texture (Information)}, and is further supported by the substantial performance gains observed when structural features are incorporated, as demonstrated in Table~\ref{tab-feature-ablation} and Table~\ref{tab-Reranker-ablation}. For the Amazon dataset, textual matching accounts for 52\% of the Top-20 candidates and contributes more answers than structural traversal. This aligns with the rich textual content observed in Amazon, as shown in Figure~\ref{fig-top20}. Overall, these results demonstrate that MoR exhibits adaptive retrieval behavior across datasets, effectively prioritizing the most informative retrieval strategy based on the underlying data characteristics. This provides strong evidence that MoR can dynamically coordinate between structural and textual retrieval sources.

\newpage
\subsection{Efficiency and Scalability of MoR}

\begin{table}[t!]
\centering
\scriptsize
\begin{tabular}{lccc}
\toprule
\textbf{Method} & \textbf{Planning} & \textbf{Reasoning} & \textbf{Organizing} \\
\midrule
\textbf{Theoretical} & $\mathcal{O}(K \cdot D)$ & $\mathcal{O}(Ad^{L-1})$ & $\mathcal{O}(Bd^{L-1})$ \\
\textbf{Empirical (s)} & 0.971 & 0.463 & 0.0340 \\
\textbf{Empirical+Batch (s)} & -- & -- & 0.0134 \\
\textbf{Empirical+Parallel (s)} & -- & 0.398 & -- \\
\bottomrule
\end{tabular}
\caption{Theoretical and empirical time complexity for each component during single-query retrieval. Batch and parallel optimization are applied to the Organizing and Reasoning stages, respectively.}
\label{tab-time-complexity}
\end{table}

\begin{table}[t!]
\centering
\vspace{-2ex}
\setlength\tabcolsep{13pt}
\small
\begin{tabular}{lccc}
\toprule
\textbf{Dataset} & \textbf{1-hop} & \textbf{2-hop} & \textbf{3-hop} \\
\midrule
\textbf{Prime}   & 0.059 s & 0.068 s & 0.072 s \\
\textbf{Amazo}n  & 0.381 s & 0.403 s & 0.669 s \\
\textbf{MAG}     & 0.289 s & 0.297 s & 1.233 s \\
\bottomrule

\end{tabular}
\caption{Empirical traversal time across different hop depths for each dataset.}
\label{tab-hop-times}
\vspace{-4ex}
\end{table}
Since MoR consists of three stages, we conduct the complexity analysis by analyzing each of these stages theoretically and empirically. The results are shown in Table~\ref{tab-time-complexity}, where $K$ means the number of tokens in the generated planning graph, $D$ is the model dimension following the GPT-style decoding and key-value caching, $d$ denotes the average degree of node at each step, $L$ means the number of steps/layers in one path, and both $A$ and $B$ are constant. Due to the limitations of LLMs, it is hard to improve the efficiency of \textbf{Planning}. As the traversal is independent, we have implemented the parallel version to speed up \textbf{Reasoning}. For the \textbf{Organizing} stage, we can easily use batch to compute the ranking scores for multiple candidates and queries simultaneously, improving the efficiency.  

Based on the above analysis, the time complexity for the whole framework is $\mathcal{O}(K\cdot D + (A+B)d^{L-1})\approx\mathcal{O}(d^{L-1})$, indicating that the complexity exponentially grows as the length of the reasoning path increases. To examine how time complexity varies with path length, we empirically analyze the scalability of MoR with respect to the depth of reasoning paths. We group test queries by the number of reasoning hops (e.g., 1-hop, 2-hop, 3-hop) and compare efficiency across these three groups on three datasets. As shown in Table~\ref {tab-hop-times}, most queries can be processed within 1s. While deeper queries naturally require more processing time, MoR maintains significantly high efficiency due to its step-wise traversal and adaptive fallback to textual retrieval when necessary. 

More detailed analysis of computational complexity and scalability to large datasets can be found in Appendix~\ref{app-complexity} and Appendix~\ref{app-scalability}.

%% file: RelatedWork.tex
\newpage
\section{Related Work}\label{sec-relatedwork}
\textbf{Retrieval-augmented Generation (RAG):}
RAG enhances generative tasks by retrieving relevant information from external knowledge sources~\cite{he2024g, gao2023retrieval} and has been widely used to improve question-answering ~\cite{liu2023knowledge}. With LLMs, RAG has been used for mitigating hallucinations~\cite{yao2023llm}, enhancing interpretability~\cite{gao2023chat}, and enabling dynamic knowledge updates~\cite{wang2024knowledge}. This work essentially leverages the idea of RAG to retrieve supporting entities from TG-KBs to contextualize answer generation. Depending on concrete types of knowledge being retrieved, existing retrievers can be categorized into structural and textual retrieval, which are reviewed next.

\noindent \textbf{Textual and Structural Retrieval:} Early textual retrieval models, such as TF-IDF and BM25~\cite{robertson2009probabilistic}, rely on lexical similarity and keyword matching~\cite{chen-etal-2017-reading,yang2019end,mao-etal-2021-generation}. Modern approaches address this limitation by learning dense representations~\cite{karpukhin-etal-2020-dense}. Beyond textual retrieval, structural retrieval leverages graph-based techniques to extract structured knowledge. Methods such as graph traversal~\cite{wang2024knowledge, Jiang2023StructGPTAG}, community detection~\cite{Edge2024FromLT}, and graph machine learning models, including graph neural networks~\cite{yasunaga2021qagnn, Mavromatis2024GNNRAGGN}, play a crucial role in structural retrieval. Our approach integrates the strengths of both textual and structural retrieval by infusing the mixture-of-expert philosophy into retrieval design.

Due to page limitation, a comprehensive version of the related work is attached in Appendix~\ref{app-comprehensive}.

%% file: Conclusion.tex
\vspace{-1ex}
\section{Conclusion}\label{sec-conclusion}
\vspace{-1ex}
In this work, we propose a mixture of structural and textual retrieval (MoR) to adaptively retrieve structural and textual knowledge based on query desire, which first utilizes a textual graph generator to generate the planning graph, then performs a mixed traversal and conducts organizing via a structure-aware reranker to obtain final candidates. Experiments demonstrate the advantages of our MoR in harmonizing the retrieval of both textual and structural knowledge with insightful discoveries, including balancing retrieval performance across queries with different patterns and query-adaptive knowledge desire for structural/textual knowledge.

%% file: Limitations.tex
\newpage
\section{Limitations}
In this paper, we integrate a mixture of expert philosophy into retrieval design and propose a Mixture of structural-and-textual Retrieval (MoR) to adaptively retrieve textual and structural knowledge. The limitations of MoR can be categorized into two main aspects in the following:

\textbf{Lack of Domain-Specific Knowledge:} Our proposed MoR, similar to other baselines, does not exhibit significantly higher performance on PRIME than AMAZON and MAG. The reason is the lack of biomedical knowledge required to comprehend biomedical questions, extract key information, navigate relevant entities and relations, and rerank retrieved candidates. This suggests that current state-of-the-art retrieval models, even paired with LLMs' intelligence, still struggle to handle domain-specific knowledge effectively. Such limitations may extend to other specialized domains, such as finance and law. Future research could integrate domain-specific knowledge into retrieval. 

\textbf{Reranking at Every Traversal Layer:} Our current MoR adaptively routes retrieved candidates into the Top-K positions at the final layer via reranking, effectively implementing a conventional Mixture of Experts (MoE) routing mechanism. Despite the state-of-the-art performance we have achieved in Table~\ref{tab-merged}, this routing mechanism could also be applied to intermediate layers, where after each retrieval step, candidates are reranked, and only Top-K proceeds to the next round of traversal and retrieval. This enables every layer of mixed traversal to emulate the router design of the MoE.

\textbf{Multi-Trajectory Reranking:} While our current Structural Reranker is designed to compute ranking scores by leveraging the full spectrum of trajectory information from multiple traversed paths ending at each candidate (as illustrated in Figure~\ref{fig-framework}), our implementation currently utilizes only the most informative trajectory (i.e., the one with the longest traversed path) due to implementation complexity. Future work should explore adaptive methods to fully integrate the complete set of traversed paths into the candidate ranking process and compare the effectiveness of leveraging traversed paths at different levels.

%% file: Appendix.tex
\section{Summary of Notations}\label{sec-notation}
\vspace{-2ex}
\begin{table}[H]
\tiny
\setlength{\extrarowheight}{.095pt}
\setlength\tabcolsep{0.5pt}
\caption{Notations and the corresponding descriptions.}
\vspace{-1ex}
\label{tab-symbols}
\begin{tabularx}{0.49\textwidth}{c|c}
\toprule
\textbf{Notations}       & \textbf{Definitions or Descriptions} \\
\midrule
$\mathcal{B}$ & Text-rich Graph Knowledge Base (TG-KB)\\

$\mathcal{V, E, D}$ & Set of Nodes, Categories and Documents of TG-KB\\

$\mathcal{D}_v, \mathcal{E}_v$ & Document and Category of Node $v$\\

$Q \in \mathcal{Q}$ & Query $Q$ from Query set $\mathcal{Q}$\\

$\mathcal{Q}^{\text{Struct}}, \mathcal{Q}^{\text{Text}}$ & Query targeted by structural and textual retrieval\\

$G = \{\mathcal{P}_i\}_{i=1}^{|G|}$ & Planning Graph consisting of multiple reasoning paths\\

$\mathcal{P}_i = (p_{i1}\rightarrow ... \rightarrow p_{iL_i})$ & Reasoning path consisting of $L_i$ sequential entities \\

$\mathcal{E}_{p_{ij}}, \mathcal{T}_{p_{ij}}$ & Textual category and restriction of path entity $p_{ij}$ \\

$\widetilde{\mathcal{C}}$ & Retrieved candidates after reasoning module.\\
\midrule
$\widetilde{\mathcal{C}}^l_i = \widetilde{\mathcal{C}}^{l, \text{Struct}}_i\cup \widetilde{\mathcal{C}}^{l, \text{Text}}_i$ & \makecell[c]{Retrieved candidates at $l^{\text{th}}$ layer for $i^{\text{th}}$ path including \\structurally retrieved ones and textually retrieved ones.}\\
\midrule

$\mathcal{C}$ & Final retrieved candidates after organizing module.\\

$P_{\mathbb{Q}\times \mathbb{G}}$ & Joint distribution of query and planning graph.\\

$\mathcal{N}_v$ & Neighborhood of entity $v$\\

$\mathcal{I}_{p_{il}}$ & Traversal Identifier of Structural and Textual Retrieval\\

$P_{\boldsymbol{\Theta}_1}$ & Planning module with its parameters $\boldsymbol{\Theta}_{1}$\\
$P_{\boldsymbol{\Theta}_2}$ & Reasoning module with its parameters $\boldsymbol{\Theta}_{2}$\\
$P_{\boldsymbol{\Theta}_3}$ & Organizing module with its parameters $\boldsymbol{\Theta}_{3}$\\
\bottomrule
\end{tabularx}
\end{table}

\begin{table}[h]
    \centering
    \scriptsize
    \begin{tabular}{lcccc}
        \toprule
        \textbf{Dataset} & \# \textbf{Entities} & \# \textbf{Text Tokens} & \# \textbf{Relations} & \textbf{Avg. Degree} \\
        \midrule
        \textbf{AMAZON} & 1,035,542 & 592,067,882 & 9,443,802 & 18.2 \\
        \textbf{MAG} & 1,872,968 & 212,602,571 & 39,802,116 & 43.5 \\
        \textbf{PRIME} & 129,375 & 31,844,769 & 8,100,498 & 125.2 \\
        \bottomrule
    \end{tabular}
    \caption{Statistics of text-rich graph knowledge bases in STaRK benchmark~\cite{wu2024stark}.}
    \label{tab-stark_stats}
\end{table}

\section{Experimental Details}\label{app-expr-setting}

\subsection{Datasets}
To evaluate the effectiveness of our proposed framework, we conduct experiments using three Text-rich Graph Knowledge Bases (TG-KBs) from STaRK~\cite{wu2024stark}. These TG-KBs cover a wide range of domains, including product reviews (Amazon), academic papers (MAG), and biomedical knowledge (Prime). Each TG-KB comprises a textual graph and an associated corpus, with the corpus containing documents linked to the nodes in the graph. Queries are meticulously crafted for each TG-KB and encompass varying levels of complexity, which desire different levels of textual and structural knowledge to answer.

\textbf{Amazon:} a dataset provides a realistic simulation of product search and recommendation. Its textual graph consists of four categories of nodes: \textit{product}, \textit{category}, \textit{color}, and \textit{brand}. Nodes are interconnected through relations such as \textit{has\_brand} and \textit{has\_category}. Textual documents encapsulate properties of corresponding nodes, such as product descriptions and customer reviews. 

\textbf{MAG:} a comprehensive resource for academic paper retrieval. In the textual graph, \textit{papers} can be connected to other nodes, such as \textit{field\_of\_study} via the \textit{paper\_has\_topic\_field\_of\_study} relation and \textit{institution} through a combination of relations like \textit{author\_affiliated\_with\_institution} and \textit{author\_writes\_paper}. Each \textit{paper} document includes the title, abstract, and metadata, such as the publication date and venue, providing rich contextual knowledge for retrieval and analysis. 

\textbf{Prime:} a highly domain-specific dataset. It focuses on medical inquiries and is sourced from the PrimeKG knowledge graph~\cite{chandak2023building}, which comprises ten entity types and eighteen relation types, offering multiple target node categories, such as \textit{disease}, \textit{gene/protein}, and \textit{drug}. The associated documents are aggregated from various databases, providing a rich and diverse source of medical knowledge.

Detailed dataset statistics are in Table~\ref{tab-stark_stats}.

\subsection{Implementation Details}\label{app-implementation}

\textbf{Planning Graph Generation:}
In Section~\ref{sec-planning}, we follow previous works~\cite{luo2023reasoning, wu2024stark} to linearize the planning process by decomposing the planning graph into sequential reasoning paths, which can be generated by LLMs via next token prediction. Given the lack of ground-truth planning graphs for training, we prompt LLMs to synthesize these ground-truth planning graphs due to their superior reasoning capability. Since the nodes in planning graphs are entity categories/types, we include \textbf{Entity Type List} in the prompt.
We also leverage in-context learning to help LLMs learn the mapping between the query and the corresponding planning graph. 
The detailed prompt for generating ground-truth planning graphs and parameters for fine-tuning the planning graph generator (i.e., Llama 3.2-3B) are shown in Prompt 1 and Table~\ref{tab-training-config}.

\begin{tcolorbox}[
colback=blue!10!white, 
colframe=blue!80!black, 
title=Prompt 1: Planning Graph Generation, 
boxsep=0.75mm, 
left=0.75mm, 
right=0.75mm, 
top=0.75mm, 
bottom=0.75mm, 
float=htbp!,     
floatplacement=tbp, 
]
\small
\label{tab- prompt}
\textbf{System Message:} You are a planning graph finder agent. Your role is to:
\newline
1. Identify the underlying **meta-path** from a given question, which consists of the **entity types** at each reasoning step. 
\newline
2. Extract the **content restriction** for each **entity type** based on the question. If there is no restriction for an entity type, leave its value empty.
\newline
You will be provided with a predefined **Entity Type List**. Only use the entity types from this list when constructing the meta-path and restrictions. Your response must be concise and strictly adhere to the specified **output format**.
\newline
\newline
\textbf{Entity Type List:} 
\textcolor{blue}{Provide the entity type list.}
\newline
\textbf{Demonstrations:} 
\textcolor{blue}{Examples for in-context learning.}
\newline
\textbf{Output Fromat:} 
\textcolor{blue}{\textit{Metapath: ""}, \textit{Restriction: \{\}}}.

\end{tcolorbox}

\begin{table}[t!]
\centering
\large
\resizebox{0.48\textwidth}{!}{
\begin{tabularx}{\textwidth}{l l X}
\toprule
\textbf{Parameter} & \textbf{Value} & \textbf{Description} \\
\midrule
\texttt{per\_device\_train\_batch\_size} & 4 & Number of training samples per device. \\
\texttt{gradient\_accumulation\_steps} & 8 & Accumulates gradients over 8 steps. \\
\texttt{num\_train\_epochs} & 100 & Number of full passes on training dataset. \\
\texttt{max\_steps} & 1000 & Maximum number of training steps; training stops once reached. \\
\texttt{learning\_rate} & 2e-4 & Initial learning rate for the optimizer. \\
\texttt{warmup\_steps} & 5 & Number of warm-up steps. \\
\texttt{optim} & adamw\_8bit & 8-bit AdamW for efficient training. \\
\texttt{lr\_scheduler\_type} & linear & Linear decay of learning rate. \\
\texttt{weight\_decay} & 0.01 & L2 regularization to prevent overfitting. \\
\texttt{seed} & 3407 & Random seed for reproducibility. \\
\bottomrule
\end{tabularx}
}
\vspace{-1ex}
\caption{Hyperparameter for planning graph generator.}
\label{tab-training-config}
\vspace{-4ex}
\end{table}

\textbf{Trajectory Collection:} As mentioned in Section~\ref{sec-organizing}, our reranker reorders the intermediate retrieved candidates based on their trajectory. To achieve this, we collect three key features: \textbf{Textual Fingerprint (TF)}, \textbf{Structural Fingerprint (SF)}, and \textbf{Traversal Identifier (TI)}. 

\textbf{Textual Fingerprint (TF):} We record the BM25 similarity scores between the query and the traversed nodes computed. Since empirical observations indicate that the length of reasoning paths is typically less than three, we fix the textual fingerprint to the length of three by padding additional 0 similarity scores for those reasoning paths whose length is less than three, allowing for batch-wise training. Additionally, we append the initial semantic ranking score of the candidate computed using cosine similarity coupled with Ada-002 embedding to the end of three BM25-based similarity scores to complement the lexical perspective. This vector is then passed through a linear layer to be transformed into an embedding of size 256. Note that this initial ranking score is also used to select the intermediate retrieved candidates used for reranking. 

\textbf{Structural Fingerprint (SF):} We concatenate the categories of all nodes in the corresponding reasoning path as a text sequence. If the reasoning path is shorter than three nodes, we prepend the sequence with "padding" tokens to ensure a fixed length. The structural fingerprint is then processed using a transformer model, which converts the sequence into an embedding of size 768, followed by a linear layer that projects it down to size 256. 

\textbf{Traversal Identifier (TI):} We track whether each node is retrieved via textual matching or structural traversal and encoding them with distinct values by initializing a learnable embedding matrix mapping each traversal identifier encoding to a 3x256-dimensional embedding vector.

After obtaining all above three trajectory features, we concatenate their obtained vectors into a unified vector (256 + 256 + 256x3 = 1280) and apply two fully connected layers to transform the combined representation into a reranking score. This score determines the final ranking.

\section{Additional Results}\label{app-results}
\subsection{Planning Performance}
To investigate the performance of our planning graph generator, we conduct an experiment to evaluate the quality of the generated planning graphs for queries. We view the planning graph as correct if it enables the retrieval of at least one ground-truth candidate by traversal following its structure. In Table~\ref{tab-planning-acc}, most planning graphs are correct on MAG and Amazon datasets, demonstrating their effectiveness in guiding the traversal. Even for the domain-specific Prime dataset, more than 60\% of planning graphs lead to correct retrieval, showing that LLMs still generate accurate plans and effectively support downstream reasoning and reranking.

\begin{table}[h!]
\centering

\begin{tabular}{lccc}
\toprule
\textbf{Dataset} & \textbf{MAG} & \textbf{Amazon} & \textbf{Prime} \\
\midrule
\textbf{Accuracy (\%)} & 88.85 & 70.03 & 60.44 \\
\bottomrule
\end{tabular}
\caption{Performance of planning graph generator on three datasets.}
\label{tab-planning-acc}
\end{table}



\subsection{Query Pattern Analysis}
\begin{figure}[h!]
    \centering
    \includegraphics[width=0.48\textwidth]{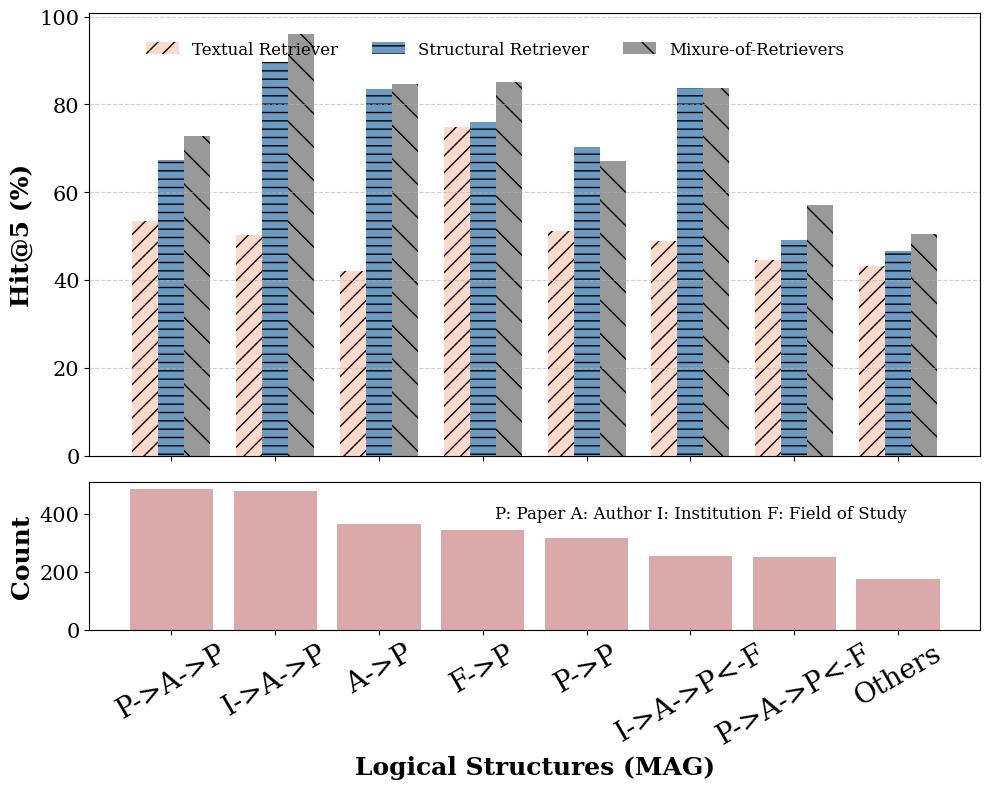}
    \caption{Imbalance number of queries and performance of different retrievers across different logic patterns.}
    \label{fig-analysis-mag}
\end{figure}
Figure~\ref{fig-analysis-mag} illustrates the analysis of query patterns in the MAG dataset. With richer relational information, queries in this dataset form a wider variety of patterns, including longer and more diverse structures. Similar to the Amazon dataset, we observe a general trend where the performance of MoR declines as the query count decreases across different patterns. Beyond this overall trend, certain query patterns in the MAG dataset stand out, such as "P → A → P" (Product-to-Author-to-Product) and "P → P" (Paper-to-Paper). Despite their relatively high occurrence, MoR still performs worse on these patterns. This is similar to low performance on the "Product → Product" pattern observed in the Amazon dataset, where repeated entity types appear within a single query. Such repetition causes the textual retriever to shift focus from the target to the repeated entities, leading to lower performance.

\section{Efficiency Analysis}
\subsection{Computational Complexity of MoR}\label{app-complexity}
Since our Planning-Reasoning-Organizing framework is multi-staged, we theoretically/empirically analyze the time complexity of each component:
\begin{itemize}[leftmargin=*]
    \item \textbf{Planning:} the planning component generates the textual planning graph by linearized token generation using a pre-trained language model, which is the well-established technique for most LLMs. Since our planning graph is category-based and its textual description consists of simple terms extracted directly from the query, the resulting textual graph is relatively simple and maintains a short token length. Therefore, the generation is highly efficient and the time complexity is linear in the number of tokens, i.e., $\mathcal{O}(K\cdot D)$ with $K$ as the number of planning graph tokens and $D$ as the model dimension, following GPT-style decoding and key-value caching.
    \item \textbf{Reasoning:} the reasoning component conducts layer-wise breadth-first traversal following generated textual planning graphs, the time complexity of which grows exponentially with the layers/hops of traversal. Let $L$ be the maximum number of layers/hops across all reasoning paths, $c$ be the total number of distinct reasoning paths decomposed from each planning graph, $d$ be the average node degree, and $t$ denote the number of nodes retrieved via textual similarity at each layer. We begin our mixed traversal from a set of $s$ seed nodes and perform layerwise expansion by retrieving neighbors of nodes from the previous layer (constrained by the prescribed node category at that corresponding layer in the planning graph). From the second layer onward, we additionally incorporate the top-$t$ nodes ranked by textual similarity to the original query and these textually retrieved nodes also contribute neighbor expansion in the next layer. Assuming $z$ represents the time used for textual matching and node entity constraint checking, the total time complexity for reasoning by following $c$ reasoning paths is 
    \begin{equation}
    \begin{aligned}
        &\mathcal{O}(c(\underbrace{(((sd + t)d + t)...)d+t}_{L})z) \\
        &=\mathcal{O}(csd^{L-1}z + ct\frac{d^{L-1}-1}{d-1}z) \\
        &\approx\mathcal{O}(Ad^{L-1})
    \end{aligned}
    \end{equation}
     as $s,t,c,z$ are typically constants with small values compared with the exponentially growing paths and are absorbed together into $A$.
    \item \textbf{Organizing:} For each candidate given by the retrieved path from reasoning stage, the organization module calculates the ranking score, which takes $\mathcal{O}(Bd^{L-1})$ with $d^{L-1}$ denotes the total number of reasoning paths and $B$ denotes the time for calculating ranking score with the defined neural network for each path.
\end{itemize}
Taking the above analysis together, the time complexity of the entire framework is $\mathcal{O}(KD + (A+B)d^{L-1})\approx\mathcal{O}(d^{L-1})$. Since our planning graph generation follows the typical LLM decoding mechanism, it does not introduce additional time overhead. Therefore, we do not focus on optimizing this stage. Instead, we optimize the Reasoning and Organizing stages, considering the exponentially growing number of retrieved paths:
\begin{itemize}[leftmargin=*]
    \item \textbf{Reasoning:} we parallelize the traversal process, as each reasoning path is explored independently. 
    \item \textbf{Organizing:} we simultaneously evaluate a batch of reasoning paths and their candidates.
\end{itemize}
We report the theoretically analyzed and empirically verified time complexity for one query retrieval in Table~\ref{tab-time-complexity}. To account for differences across datasets and reduce the impact of stochasticity in the experiments, we perform three runs per dataset and report the average efficiency. We can see that despite the exponentially growing time complexity of reasoning and organizing, the empirical time consumption is rather low.

\subsection{Scalability of Mixed Traversal to Large-scale TG-KBs}\label{app-scalability}

Our reasoning process conducts layer-wise breadth-first traversal following generated textual planning graphs, the time complexity of which grows exponentially with the increase of traversal depth. The above analysis shows that the time cost (i.e., $\mathcal{O}(Ad^{L-1})$) is mainly governed by the average node degree $d$ and the reasoning hop $L$. Therefore, even for large-scale TG-KBs, the traversal remains efficient since most graphs are sparse and the reasoning paths are reasonably short, both of which have been empirically verified in ~\cite{wu2024stark}. 

Moreover, the three TG-KBs in our experiments are already large in scale while our reasoning traversal method achieves significant efficiency in addressing each query (see Table~\ref{tab-stark_stats} and Table~\ref{tab-scala}). Furthermore, thanks to the highly parallelizable nature of our traversal process, this runtime can be further reduced by leveraging multi-processing over traversing multiple reasoning paths.

\begin{table}[t!]
\centering
\small

\setlength\tabcolsep{15pt}
\begin{tabular}{lcc}
\hline
\textbf{Dataset} & \textbf{Sequential (s)} & \textbf{Parallel (s)} \\
\hline
Amazon & 0.416 & 0.271 \\
MAG    & 0.891 & 0.846 \\
Prime  & 0.081 & 0.076 \\
\hline
\end{tabular}
\caption{Sequential vs. parallel processing time of different datasets.}
\vspace{-2ex}
\label{tab-scala}
\end{table}

\section{Extending to Question Answering Task}
Prior studies have shown that improved retrieval often leads to improved question-answering outcomes~\cite{mao2020generation,lewis2020retrieval}. To verify this positive correlation, we additionally evaluate the QA performance based on the retrieved information. Specifically, we use LLM GPT-4o as a judge to assess which retrieval method provides more helpful context for answering the question, allowing us to examine whether improvements in retrieval quality can convert to better QA performance. To mitigate hallucination, we enable LLM to respond with "I don't know" when neither retrieved candidate supports an answer. We sample 100 queries for each of the three TG-KBs and retrieve top-5 candidates by using both our proposed MoR and the classical textual matching baseline BM25. To mitigate the impact of positional bias, we evaluate two different presenting orders for candidates retrieved by the two methods, resulting in 200 evaluations per dataset. The win ratios (i.e., the percentage of preferences) for both methods across three datasets are reported in Table~\ref{tab-qa}.

\begin{table}[t!]
\centering
\setlength\tabcolsep{20pt}
\begin{tabular}{lcc}
\toprule
\textbf{Dataset} & \textbf{BM25} & \textbf{MoR} \\
\midrule
MAG     & 27.0\%   & 64.0\% \\
Prime   & 33.0\%   & 56.5\% \\
Amazon  & 40.0\%   & 42.5\% \\
\bottomrule
\end{tabular}
\caption{Comparison of BM25 and MoR QA performance across datasets.}
\label{tab-qa}
\end{table}

\section{Comprehensive Related Work}\label{app-comprehensive}
\subsection{Retrieval-augmented Generation (RAG)}
With the unprecedented success of recent LLMs in approaching human-level intelligence, retrieving relevant knowledge to support downstream generation has become increasingly crucial. Retrieval-augmented generation enhances generative tasks by integrating relevant information from external knowledge sources~\cite{he2024g, gao2023retrieval, han2024retrieval} and has been widely adopted to improve question-answering~\cite{liu2023knowledge}. In the context of LLMs, RAG has been utilized to mitigate hallucinations~\cite{yao2023llm}, enhance interpretability~\cite{gao2023chat}, and enable dynamic knowledge updates~\cite{wang2024knowledge}. This work leverages RAG to retrieve supporting entities from TG-KBs, providing contextual grounding for answer generation. Depending on the type of knowledge retrieved, existing retrievers can be classified into structural and textual retrieval approaches, which are reviewed next.

\subsection{Textual and Structural Retrieval}
Since real-world knowledge is commonly stored in both textual and structural formats~\cite{kolomiyets2011survey}, such as indexed texts and knowledge graphs, each requires a retrieval method tailored to its unique representation. Textual retriever retrieves knowledge based on its similarity to the given query and can be categorized into: lexical methods (e.g., TF-IDF and BM25~\cite{robertson2009probabilistic}) and semantic methods (e.g., DPR and Contriever~\cite{karpukhin-etal-2020-dense, izacardunsupervised}). Despite their broad applicability, the predefined linguistic rules and embedding-based semantics may struggle to capture the structural knowledge stored in graph-structured knowledge bases such as knowledge graphs and text-rich networks. To address this challenge, structural retrieval has been proposed by using graph analysis techniques (e.g., graph traversal~\cite{wang2024knowledge, Jiang2023StructGPTAG, Zhang2022GreaseLMGR, Edge2024FromLT}) and graph machine learning models (e.g., graph neural networks~\cite{yasunaga2021qagnn, Mavromatis2024GNNRAGGN}). Early methods extract local subgraphs of seeding nodes~\cite{yasunaga2021qagnn, Taunk2023GrapeQAGA} or pre-define paths approaching answers (e.g., shortest paths~\cite{luo2023reasoning, Delile2024GraphBasedRC}). To avoid exponentially expanding neighbors in the local subgraphs and break the rigid logic routined by pre-defined paths, recent advancements integrated LLMs to dynamically adjust graph traversal~\cite{Sun2023ThinkonGraphDA, wang2024knowledge, Jin2024GraphCA}. While promising, frequently invoking LLMs introduces prohibitive resource overhead. Despite the above advancements in both textual and structural retrieval, they are often applied independently and fail to mutually reinforce each other. This motivates the recent research trend of developing hybrid retrieval, which is reviewed next.

\subsection{Hybrid Retrieval}
Recently, several works have explored hybrid knowledge retrieval from TG-KBs. One approach~\cite{xia2024knowledge, li2024multi} aggregate documents from neighboring nodes, with~\citet{xia2024knowledge} applying relational filtering to remove irrelevant neighbors and~\citet{li2024multi} weighting neighbors based on field importance. Another approach~\cite{Lee2024HybGRAGHR} uses LLMs to choose either structural or textual retrieval. In contrast, our proposed MoR fully leverages the graph structure and rich texts by integrating textual matching and graph traversal into a unified framework, enabling a more seamless and interpretable interaction between structural and textual knowledge 
